%% file: main.tex
\documentclass{article}

\usepackage{arxiv}
\usepackage[utf8]{inputenc}
\usepackage[T1]{fontenc}
\usepackage{amsmath,amssymb,bm}
\input{math_commands.tex}

\usepackage[table]{xcolor}
\usepackage{graphicx}
\usepackage{float}
\usepackage{booktabs}
\usepackage{multirow}
\usepackage[round,authoryear]{natbib}
\usepackage{microtype}
\usepackage{etoolbox}
\usepackage[
    colorlinks=true,
    citecolor=blue,
    linkcolor=black,
    urlcolor=blue
]{hyperref}
\usepackage{url}

\definecolor{bestcolor}{RGB}{255,183,183}
\definecolor{secondbestcolor}{RGB}{255,242,0}
\newcommand{\best}[1]{\cellcolor{bestcolor}#1}
\newcommand{\secondbest}[1]{\cellcolor{secondbestcolor}#1}

\renewcommand{\shorttitle}{LonGaussian}
\renewcommand{\headeright}{Preprint}
\renewcommand{\undertitle}{}
\makeatletter
\patchcmd{\@maketitle}{\textsc{\undertitle}\\}{}{}%
  {\PackageError{longaussian}{Could not remove the title subtitle row}{Check arxiv.sty.}}
\patchcmd{\@maketitle}
  {\vskip 0.4in \@minus 0.1in \center{\@date} \vskip 0.2in}
  {\par\vskip 12pt}{}%
  {\PackageError{longaussian}{Could not compact the empty date area}{Check arxiv.sty.}}
\makeatother
\date{}

\title{LonGaussian: Tracking Foundation Priors for Long-Context Feedforward Gaussian Reconstruction}
\newcommand{\authorinfo}[3]{%
  \begin{minipage}[t]{0.31\linewidth}
    \centering\normalfont
    {\large #1\par}
    \small
    #2\\
    Huazhong University of Science and\\
    Technology\\
    Wuhan, Hubei, China\\
    \texttt{#3}
  \end{minipage}%
}
\newcommand{\longaussianauthors}{%
  \begin{minipage}[t]{\dimexpr\textwidth-2\tabcolsep\relax}
    \centering\normalfont
    \authorinfo{Zhoufeng Xie}{School of SE}{zhoufengxie@hust.edu.cn}\hfill
    \authorinfo{Guanjun Wu}{School of CS}{guajuwu@hust.edu.cn}\hfill
    \authorinfo{Shuai Zhang}{School of EIC}{shuaizhang@hust.edu.cn}
    \par\vspace{10pt}
    \authorinfo{Xinggang Wang}{School of EIC}{xgwang@hust.edu.cn}%
    \hspace{0.08\linewidth}%
    \authorinfo{Wenyu Liu\thanks{Corresponding author.}}{School of EIC}{liuwy@hust.edu.cn}
    \par\vspace{10pt}
    {\small\url{https://abearkeeper.github.io/LonGaussian.github.io/}}
  \end{minipage}%
}
\author{\longaussianauthors}

\hypersetup{
    pdftitle={LonGaussian: Tracking Foundation Priors for Long-Context Feedforward Gaussian Reconstruction},
    pdfauthor={Zhoufeng Xie, Guanjun Wu, Shuai Zhang, Xinggang Wang, Wenyu Liu},
    pdfsubject={Long-context feed-forward 3D Gaussian reconstruction},
    pdfkeywords={3D Gaussian Splatting, feed-forward reconstruction, long-context reconstruction, point tracking},
    bookmarksnumbered=true
}

\begin{document}
\maketitle

\begingroup
\setlength{\intextsep}{0pt}
\begin{figure}[H]
    \centering
    \includegraphics[width=0.76\textwidth]{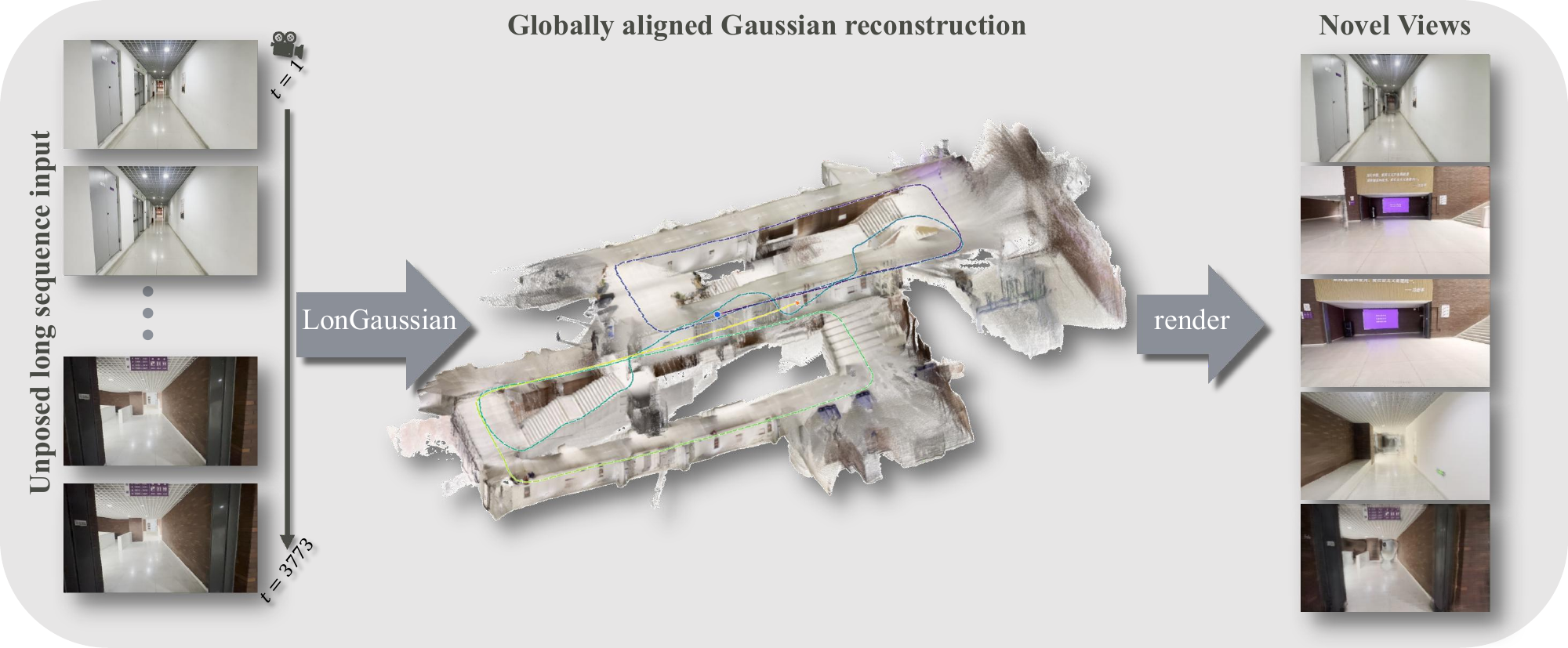}

    \caption{Ultra-long reconstruction with known intrinsics and unknown camera poses. LonGaussian accumulates Gaussians and camera poses in a shared coordinate system to support novel-view rendering.}
    \label{fig:applications}

\end{figure}
\endgroup

\begin{abstract}
Reconstructing high-quality, renderable 3D scenes from long videos is essential for robotic perception and embodied intelligence. Many feed-forward Gaussian reconstruction methods rapidly predict 3D scenes from a small set of images, but extending them to longer sequences presents two bottlenecks: jointly processing all views increases GPU memory requirements, while accumulating pixel-aligned predictions redundantly represents the same surfaces and increases storage and rendering costs. We present \textbf{LonGaussian}, a streaming feed-forward framework that reconstructs compact, globally renderable Gaussian scenes from long image sequences with known intrinsics and unknown camera poses. Built on a pretrained streaming 3D reconstruction backbone, LonGaussian decodes long-context geometric features into dense Gaussians. Within each chunk, multi-frame point tracks link repeated observations of the same surface and guide a learned merger to consolidate the corresponding Gaussian predictions. Rendered-depth consistency filtering further suppresses Gaussians from other chunks that incorrectly occlude locally reconstructed surfaces. On DL3DV and RealEstate10K benchmarks, LonGaussian improves novel-view rendering quality while using substantially fewer Gaussians than the evaluated feed-forward baselines.
\end{abstract}

\section{Introduction}
\label{sec:introduction}

Reconstructing renderable 3D scenes from long videos supports applications in robotic perception, embodied intelligence, and immersive media. 3D Gaussian Splatting (3DGS) represents scenes with explicit Gaussian primitives, enabling high-quality novel-view synthesis and real-time rendering~\citep{kerbl20233d}. Optimization-based reconstruction methods produce high-quality scene representations but require time-consuming per-scene iterative optimization~\citep{mildenhall2021nerf,kerbl20233d,huang20242d,Wu_2024_CVPR,zhang2025dynamic}. Feed-forward Gaussian reconstruction improves efficiency by directly predicting renderable 3D representations from multiple images through network inference~\citep{charatan2024pixelsplat,chen2024mvsplat}. However, many existing models jointly process the input views~\citep{charatan2024pixelsplat,chen2024mvsplat,xu2025depthsplat,jiang2025anysplat}, causing computational and GPU memory requirements to grow rapidly with sequence length. These demands limit long-video inference, whose memory requirements may exceed GPU capacity.

\begin{figure}[t]
    \centering
    \includegraphics[width=0.86\textwidth]{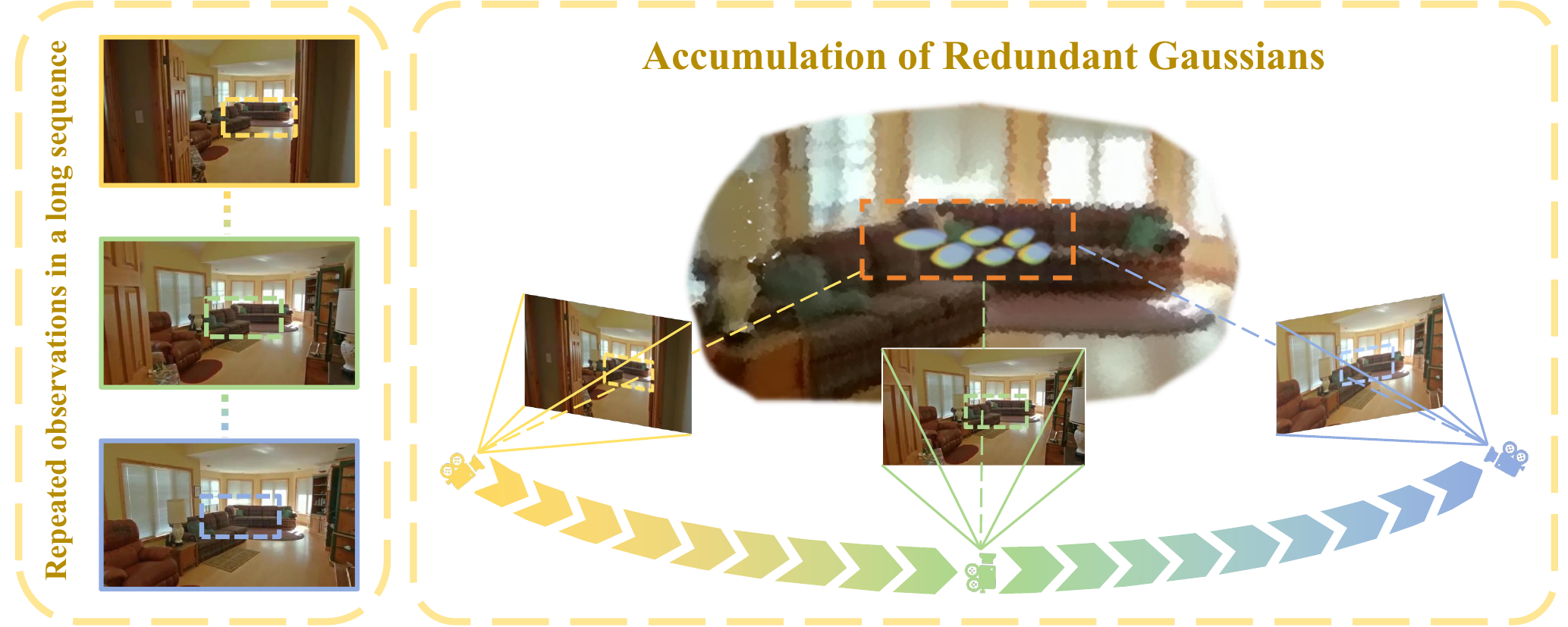}

    \caption{Gaussian redundancy from repeated observations. Pixel-aligned predictions accumulate overlapping primitives for the same surface, increasing storage and rendering costs; matching colors denote corresponding observations and Gaussians.}
    \label{fig:redundancy_problem}

\end{figure}

Scaling to long sequences is constrained by both the inference backbone and the explicit scene representation. Online approaches alleviate joint-view computational bottlenecks by updating an accumulated Gaussian scene or a compressed historical state~\citep{chen2026of3gsontheflyfeedforward3d,cheng2026recosplat,song2026streamsplat,huang2026longsplat}. However, many feed-forward Gaussian predictors still associate one primitive with each input pixel. Accumulating predictions from repeated surface observations retains overlapping Gaussians for the same geometry, increasing representation size and rasterization workload with sequence length. Figure~\ref{fig:redundancy_problem} illustrates this redundancy. Although LoGeR enables streaming long-context geometric reconstruction through chunk-wise processing and hybrid memory~\citep{zhang2026loger}, it predicts dense geometry and camera trajectories rather than maintaining a compact renderable scene. Extending feed-forward reconstruction to long videos therefore requires both streaming geometric processing and explicit control of redundancy caused by repeated observations.

Tracking foundation models offer a natural prior for addressing this redundancy. A point trajectory links image observations of the same physical surface across frames and records surface visibility~\citep{harley2022particle,doersch2023tapir,karaev2024cotracker,karaev2025cotracker3,xiao2024spatialtracker,harley2025alltracker}. These trajectories can guide Gaussian grouping across repeated observations. In contrast to spatial strategies that aggregate primitives according to voxel occupancy or Euclidean proximity, track-based grouping uses explicit cross-frame correspondences, providing a more direct and interpretable basis for consolidation.

To this end, we propose \textbf{LonGaussian}, a streaming feed-forward Gaussian reconstruction framework for long image sequences. LoGeR provides the backbone for predicting renderable 3D Gaussians, using chunk-wise hybrid memory to process local chunks while retaining long-range geometric context. An independent tracker within each chunk groups corresponding pixels into multi-frame trajectories. A learned merger consolidates their Gaussian predictions into representative primitives, reducing redundancy while integrating geometric and appearance evidence. Rendered-depth consistency filtering further suppresses residual cross-chunk floaters. Together, these components reconstruct compact, globally renderable scenes from long sequences while preserving detail for high-quality rendering. Figure~\ref{fig:applications} provides a qualitative example of scene reconstruction and novel-view rendering from an ultra-long unposed sequence.

Our contributions are summarized as follows:
\begin{itemize}
    \item We propose \textbf{LonGaussian}, a streaming feed-forward framework that extends LoGeR's long-context geometric representation to the reconstruction of compact, globally renderable Gaussian scenes.

    \item We introduce a Gaussian consolidation mechanism that uses explicit cross-frame correspondences from multi-frame tracks to merge redundant predictions within each chunk, together with rendered-depth consistency filtering for residual cross-chunk artifacts.

    \item Experiments on DL3DV and RealEstate10K demonstrate high-quality novel-view synthesis with substantially fewer Gaussians than existing feed-forward approaches; an ultra-long real-world sequence further demonstrates qualitative applicability at scale.
\end{itemize}

\section{Related Work}
\label{sec:related_work}

\subsection{Feed-Forward and Long-Context 3D Reconstruction}

Feed-forward 3D reconstruction has progressed from pairwise point-map regression and matching models such as DUSt3R and MASt3R~\citep{wang2024dust3r,leroy2024grounding} to architectures that process larger image collections or streams. Offline models such as VGGT and $\pi^3$ jointly reason over all input views through global attention, whose computational and memory costs grow rapidly with sequence length~\citep{wang2025vggt,wang2026pi,yang2025fast3r,cabon2025must3r,tang2025mv,lin2026depth,keetha2026mapanything,wang2026vggtomega}. CUT3R instead processes observations sequentially while maintaining a persistent recurrent state~\citep{wang2025continuous,wang20253d,liu2025slam3r,chen2025long3r,lan2026stream3r}. TTT3R reinterprets CUT3R's recurrent update as test-time training and derives a confidence-aware learning rate to improve generalization to longer sequences~\citep{chen2026ttt3r}. Separately, LaCT improves the hardware efficiency and state capacity of test-time training by updating fast weights over large token chunks~\citep{zhang2026test,Elflein_2026_CVPR,jin2026zipmap}. LoGeR combines chunk-wise bidirectional reasoning with sliding-window attention for uncompressed local context and TTT fast weights for compressed global context~\citep{zhang2026loger}. These models primarily predict point maps, depth, and camera parameters; a suitable bridge to compact, explicit Gaussian scenes remains missing. We build on LoGeR's long-context geometry to address this complementary problem of feed-forward reconstruction of compact, globally renderable scenes.

\subsection{Feed-Forward Gaussian Reconstruction}

Feed-forward Gaussian reconstruction predicts image-conditioned 3D Gaussians from multi-view inputs~\citep{charatan2024pixelsplat,chen2024mvsplat,xu2025depthsplat,smart2024splatt3r,ye2025no,kang2025selfsplat,xu2025freesplatter,huang2025no,ye2026yonosplat}, with AnySplat further supporting uncalibrated image collections~\citep{jiang2025anysplat}. Because these predictions are commonly pixel-aligned, overlapping views may assign redundant or inconsistent primitives to the same physical surface, causing the accumulated representation to grow with sequence length~\citep{park2026ecosplat,zhang2026anchorsplat,miao2025evolsplat,xia2025csg,guo2026salon3r,smart2026learning}. Recent approaches support online or long-sequence reconstruction through recursive fusion, rendering feedback, or persistent scene memory~\citep{chen2026of3gsontheflyfeedforward3d,cheng2026recosplat,song2026streamsplat,huang2026longsplat,li2025streamgs}, but preserving geometry and rendering quality remains challenging as sequences grow. Our framework builds on LoGeR's established long-context backbone. Bidirectional within-chunk reasoning and global TTT memory establish coherent geometry before Gaussian consolidation, separating long-context inference from explicit scene maintenance.

\subsection{Compact Gaussian Representations}

Compression of optimization-based 3DGS commonly relies on pruning, quantization, codebooks, or entropy coding~\citep{lee2024compact,fan2024lightgaussian,hanson2025pup}. These methods operate on per-scene optimized representations and often need training-view statistics or refinement, complicating integration into feed-forward reconstruction without additional per-scene processing. Feed-forward approaches instead control the number of primitives through geometry-adaptive allocation~\citep{fei2024pixelgaussian,park2026ecosplat,moreau2026off}, anchor- or voxel-aligned representations~\citep{zhang2026anchorsplat,guo2026salon3r}, and matching-based fusion~\citep{xia2025csg}. Such strategies typically consolidate Gaussians according to importance, spatial proximity, or local matches~\citep{smart2026learning}, but do not explicitly organize repeated observations of the same surface through consistent multi-frame correspondences. In contrast, LonGaussian uses tracker-produced trajectories as explicit consolidation units, grouping track-supported observations across an entire chunk before fusing their Gaussian predictions. Resolving correspondence before consolidation targets the source of long-sequence redundancy rather than a post-hoc density problem, while preserving efficient feed-forward inference without per-scene refinement.

\section{Method}
\label{sec:method}

\begin{figure}[t]
    \centering
    \includegraphics[width=\textwidth]{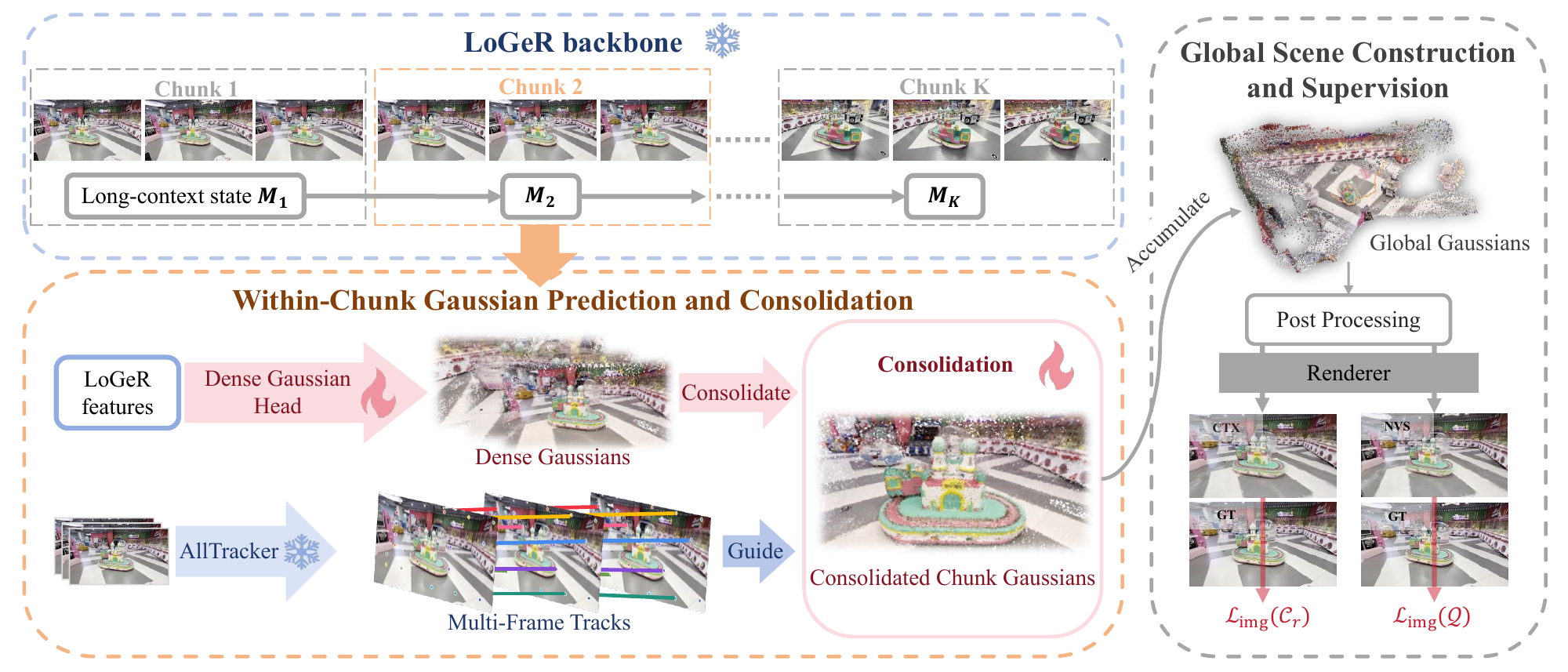}

    \caption{Overview of \textbf{LonGaussian}. LoGeR processes overlapping chunks, and AllTracker correspondences guide Gaussian consolidation before cross-chunk filtering and far-field modeling. Snowflakes and flames denote frozen and trainable modules, respectively.}
    \label{fig:framework}

\end{figure}

\subsection{Preliminaries}
\label{sec:preliminaries}

\paragraph{3D Gaussian Splatting.}
3DGS represents a scene as a set of anisotropic Gaussian primitives \(\mathcal{G}=\{g_i\}_{i=1}^{M}\)~\citep{kerbl20233d}. Each primitive is parameterized as
\begin{equation}
g_i
=
\bigl(
\boldsymbol{\mu}_i,\,
\mathbf{q}_i,\,
\mathbf{s}_i,\,
\alpha_i,\,
\boldsymbol{\theta}_i
\bigr),
\end{equation}
where \(\boldsymbol{\mu}_i\in\mathbb{R}^{3}\) is the center, \(\mathbf{q}_i\in\mathbb{R}^{4}\) is a unit quaternion encoding rotation, \(\mathbf{s}_i\in\mathbb{R}^{3}_{>0}\) is the anisotropic scale, \(\alpha_i\in(0,1)\) is the opacity, and \(\boldsymbol{\theta}_i\) contains spherical-harmonic (SH) coefficients for view-dependent color. A differentiable rasterizer \(\mathcal{R}\) renders an image from viewpoint \((\mathbf{T},\mathbf{K})\) as \(\hat{\mathbf{I}}=\mathcal{R}(\mathcal{G},\mathbf{T},\mathbf{K})\). We therefore adopt 3DGS as our scene representation: its explicit primitives from repeated observations can be consolidated directly, while differentiable rendering provides image-space supervision for the resulting global scene.

\paragraph{Long-Context Geometric Reconstruction.}
Long-Context Geometric Reconstruction (LoGeR) reconstructs long image sequences through chunk-wise processing and hybrid memory~\citep{zhang2026loger}. It partitions a sequence into overlapping chunks of length \(L\) with \(O\) shared frames and a stride of \(L-O\). Within each chunk, bidirectional interaction captures local geometry; across chunks, Sliding-Window Attention (SWA) preserves detailed recent context, while persistent TTT fast weights carry compressed long-range information. For chunk \(\mathcal I_k\) with frame-index set \(\mathcal J_k\), we summarize the backbone's prediction interface as
\begin{equation}
\left(\mathbf H_k,\{\mathbf P_t,\mathbf T_t\}_{t\in\mathcal J_k}\right)
=\mathcal B\!\left(\mathcal I_k;\mathbf M_{k-1}\right),
\label{eq:loger_backbone}
\end{equation}
where \(\mathbf H_k\) denotes long-context features, \(\mathbf P_t\) and \(\mathbf T_t\) are the dense point map and camera pose for frame \(t\), and \(\mathbf M_{k-1}\) comprises the SWA cache and TTT fast weights carried from preceding chunks. LoGeR updates memory per chunk and registers successive chunks using overlapping-frame pose estimates. LonGaussian inherits this pretrained backbone, memory, and alignment without modification. It extends LoGeR's dense geometry and camera motion to compact, globally renderable scenes by decoding long-context features into 3D Gaussians and consolidating repeated surface observations.

\subsection{LonGaussian Framework}
\label{sec:framework}

Given an unposed RGB sequence \(\mathcal I=\{\mathbf I_t\}_{t=1}^{N}\) with known intrinsics \(\mathbf K\), LonGaussian estimates camera poses and reconstructs a compact global Gaussian scene:
\begin{equation}
\mathcal G=F(\mathcal I,\mathbf K).
\label{eq:longaussian_pipeline}
\end{equation}
We decompose \(F\) into dense prediction, track-guided consolidation, and scene-level artifact handling (Figure~\ref{fig:framework}). For each overlapping chunk \(\mathcal I_k\), the frozen LoGeR backbone \(\mathcal B\) extracts geometry and features using the long-context state \(\mathbf M_{k-1}\) from preceding chunks. A trainable Gaussian branch \(\mathcal D_\psi\) decodes these outputs, and the inherited alignment operator \(\mathcal W_k\) maps the predictions to the first chunk's coordinate system:
\begin{equation}
\mathcal G_k^d=\mathcal W_k\!\left[\mathcal D_\psi\!\left(\mathcal B(\mathcal I_k,\mathbf K_k;\mathbf M_{k-1})\right)\right].
\label{eq:chunk_gaussian_prediction}
\end{equation}
Each context pixel produces one Gaussian, centered at its predicted 3D point plus a learned offset. Frozen AllTracker \(\mathcal A\) extracts multi-frame tracks directly from \(\mathcal I_k\), linking repeated surface observations. The consolidation operator \(\mathcal C_\phi\) replaces track-associated predictions with representative Gaussians while retaining uncovered predictions. After the aligned chunk representations are accumulated, the scene operator \(\mathcal S\) applies rendered-depth consistency filtering and far-field shell construction:
\begin{equation}
\mathcal G=\mathcal S\!\left(\bigcup_{k=1}^{K}
\mathcal C_\phi\!\left(\mathcal G_k^d,\mathcal A(\mathcal I_k)\right)\right).
\label{eq:global_scene_construction}
\end{equation}
Here, globally renderable denotes a single Gaussian representation in a shared coordinate system, without guaranteeing drift-free alignment on long-range revisits.

\subsection{Track-Guided Gaussian Consolidation}
\label{sec:track_consolidation}

The consolidation operator groups predictions by correspondence before merging them. For each chunk, frozen AllTracker~\citep{harley2025alltracker} associates image locations across frames. Each location indexes the corresponding pixel-aligned Gaussian \(g_{mt}\). We retain observations whose tracking confidence exceeds \(\tau_{\mathrm{trk}}\) and assign each frame--pixel prediction to at most one track. Let \(\Omega_m\) denote the retained observations of track \(m\). A track is accepted only if it contains at least \(n_{\min}\) observations, its mean opacity is at least \(\tau_\alpha\), and the root-mean-square distance of its centers from their arithmetic mean is at most \(\tau_{\mathrm{geo}}\).

The geometric acceptance test is
\begin{equation}
\left(\frac{1}{|\Omega_m|}\sum_{t\in\Omega_m}
\|\boldsymbol\mu_{mt}-\tilde{\boldsymbol\mu}_m\|_2^2\right)^{1/2}
\leq\tau_{\mathrm{geo}},
\label{eq:track_geometry}
\end{equation}
where \(\tilde{\boldsymbol\mu}_m\) is the mean center of the retained observations. This test rejects tracks whose associated predictions do not agree on a common 3D location.

For each accepted track, we construct a base Gaussian \(\bar g_m\) using opacity-weighted attribute averages, alpha compositing, and projected rotation averaging. A permutation-invariant network refines this base using pooled information from the associated observations:
\begin{equation}
\hat g_m=\bar g_m\oplus F_\phi\!\left(\operatorname{Pool}\{g_{mt}:t\in\Omega_m\}\right).
\label{eq:track_consolidation}
\end{equation}
The pooled descriptor contains the mean and standard deviation of the centers, scales, opacities, and SH coefficients, together with the visible-frame fraction, summarizing observation agreement and track support. The operator \(\oplus\) applies additive corrections to the center and SH coefficients and bounded corrections in log-scale and logit-opacity space, while retaining the base rotation. Zero-initialized output layers make the merger initially reproduce deterministic aggregation before learning residual corrections through rendering supervision. Each accepted track contributes one refined representative in place of its member Gaussians. Uncovered predictions from the chunk's non-overlapping contribution range are retained to preserve scene coverage. Together, these representatives and uncovered predictions form the output of \(\mathcal C_\phi\).

\subsection{Scene-Level Artifact Handling}
\label{sec:scene_refinement}

The scene operator \(\mathcal S\) addresses residual cross-chunk occlusions and unreliable sky geometry. To identify foreground floaters, we render each chunk's scene to obtain reference depth and opacity, then project Gaussians from other chunks into that chunk's views. Comparisons exclude neighborhoods with insufficient opacity support or excessive depth spread. For Gaussian \(i\) projected into frame \(t\) of chunk \(k\), we measure
\begin{equation}
\delta_{ikt}=\frac{\widetilde D_{kt}(\mathbf u_{ikt})-z_{ikt}}
{\widetilde D_{kt}(\mathbf u_{ikt})},
\label{eq:relative_depth_gap}
\end{equation}
where \(z_{ikt}\) is the Gaussian's camera-space depth, \(\mathbf u_{ikt}\) is its projected position, and \(\widetilde D_{kt}\) is the local median reference depth. Positive values indicate a Gaussian in front of the reference surface. A reliable comparison contributes an occlusion vote when \(\delta_{ikt}\geq\tau_{\mathrm{gap}}\), and an agreement vote when \(|\delta_{ikt}|\leq\tau_{\mathrm{agree}}\).

Removal requires at least \(n_{\mathrm{vote}}\) occlusion votes, an occlusion fraction of at least \(\rho_{\mathrm{vis}}\) among all reliable comparisons, and an occlusion fraction of at least \(\rho_{\mathrm{dec}}\) among the combined occlusion and agreement votes. This conservatively suppresses persistent foreground floaters; a chunk never votes against its own predictions. For sky pixels identified by a fixed segmenter, we replace finite-depth Gaussians with viewing-direction observations on a shell of radius \(r_{\mathrm{sky}}=\gamma_{\mathrm{sky}}r_{\mathrm{scene}}\). Angular binning yields one isotropic, degree-zero-SH Gaussian per occupied bin, compactly representing the far-field appearance.

\subsection{Training Strategy and Objectives}
\label{sec:training_objectives}

We train the Gaussian branch and merger in two stages while keeping LoGeR and AllTracker frozen. For stage \(s\in\{1,2\}\), the complete objective is
\begin{equation}
\mathcal L^{(s)}=
\mathcal L_{\mathrm{img}}(\mathcal C_r)
+\lambda_{\mathrm{tgt}}\mathcal L_{\mathrm{img}}(\mathcal Q)
+\lambda_{\mathrm{aniso}}^{(s)}\mathcal L_{\mathrm{aniso}}
+\lambda_{\mathrm{off}}^{(s)}\mathcal L_{\mathrm{off}}.
\label{eq:training_objective}
\end{equation}
The set \(\mathcal C_r\) contains a small subset of context views; target views \(\mathcal Q\) never contribute Gaussians. For either view set \(\mathcal V\), the image objective is
\begin{equation}
\mathcal L_{\mathrm{img}}(\mathcal V)=
\frac{1}{|\mathcal V|}\sum_{t\in\mathcal V}
\left(\lambda_{\mathrm{rgb}}\ell_{\mathrm{rgb}}^{t}
+\lambda_{\mathrm{lpips}}\ell_{\mathrm{lpips}}^{t}
+\lambda_{\mathrm{ssim}}\ell_{\mathrm{ssim}}^{t}\right).
\label{eq:per_view_image_loss}
\end{equation}
The three per-view terms measure RGB mean-squared error, LPIPS, and \(1-\operatorname{SSIM}\), respectively, between the rendered image and its ground truth.

Stage~I trains the dense Gaussian branch on short sequences with nonzero \(\lambda_{\mathrm{aniso}}^{(1)}\) and \(\lambda_{\mathrm{off}}^{(1)}\), so the corresponding terms regularize anisotropic scale and the learned offset from the point map. Stage~II initializes from Stage~I and freezes the feature decoder. Stage~IIa trains only the merger on mixed-length DL3DV clips; Stage~IIb then jointly adapts the merger and, on short clips only, the attribute head using DL3DV and RE10K. We set \(\lambda_{\mathrm{aniso}}^{(2)}=\lambda_{\mathrm{off}}^{(2)}=0\), so Stage~II directly supervises the consolidated scene through the context- and target-view image losses in Eq.~\eqref{eq:training_objective}. The final training stage includes depth filtering and the far-field shell. Correspondence, ownership, filtering, and sky assignment are treated as non-differentiable routing decisions; gradients pass through the retained learned Gaussian attributes and differentiable rendering.

\section{Experiments}
\label{sec:experiments}

\subsection{Experimental Setup}

\paragraph{Baselines.}
We compare LonGaussian with the offline methods AnySplat~\citep{jiang2025anysplat} and WorldMirror~\citep{liu2026worldmirror}, and the streaming methods OF3GS~\citep{chen2026of3gsontheflyfeedforward3d} and ReCoSplat~\citep{cheng2026recosplat}. For ReCoSplat, we report both its released low-resolution checkpoint and a locally resolution-adapted checkpoint; the latter is an accommodation for our evaluation resolution rather than an official training reproduction. StreamSplat~\citep{song2026streamsplat}, SaLon3R~\citep{guo2026salon3r}, and StreamGS~\citep{li2025streamgs} are omitted because runnable official implementations were unavailable at the time of evaluation.

\paragraph{Datasets and Metrics.}
We train on DL3DV and use 150 sequences from the official RE10K training split for final adaptation. We evaluate on DL3DV-140 and 180 eligible sequences from the official RE10K test split; identity-level audits confirm zero overlap between training and evaluation sequences. Images are resized to $280\times490$; each scene is constructed from up to 128 context frames, and held-out views serve as targets. LonGaussian uses two-pass unposed evaluation: the first pass processes context and target images to estimate camera poses, while the second resets the TTT states and SWA cache and constructs the scene from context images only. We align the first-pass trajectory to the second-pass scene using shared context cameras and render held-out views at the aligned target poses (Appendix~\ref{app:two_pass_evaluation}). Methods share frame indices whenever their interfaces permit, and ReCoSplat retains its prescribed first-chunk scale matching. We average PSNR, SSIM, LPIPS, and the final Gaussian count over all held-out views and scenes.

\paragraph{Implementation Details.}
We initialize and freeze the public LoGeR backbone and set $L=12$, $O=3$. With a global batch size of four, Stage~IIa trains only the merger on DL3DV for 900 steps; Stage~IIb adapts for 300 steps, updating the attribute head only at 16 frames and the merger at every length. Both uniformly sample $|\mathcal C|\in\{16,64,96,128\}$. Training uses AdamW on 24-GB RTX 3090 GPUs. Depth filtering and the far-field shell are enabled in final-stage training and evaluation; the appendix provides the remaining settings.

\subsection{Results}

\begin{table}[t]
\caption{Main results on DL3DV and RealEstate10K using complete reconstructed scenes for target-view rendering. \#G is the average Gaussian count in millions; \colorbox{bestcolor}{pink}/\colorbox{secondbestcolor}{yellow} mark best/second-best. ``OOM'' exceeds 24-GB GPU memory.}
\label{tab:main_results}
\centering
\scriptsize
\setlength{\tabcolsep}{2.4pt}
\renewcommand{\arraystretch}{1.05}
\resizebox{\textwidth}{!}{%
\begin{tabular}{@{}clcccccccccccc@{}}
\toprule
Dataset & Method & \multicolumn{4}{c}{64 frames} & \multicolumn{4}{c}{96 frames} & \multicolumn{4}{c}{128 frames} \\
\cmidrule(lr){3-6}\cmidrule(lr){7-10}\cmidrule(lr){11-14}
& & PSNR$\uparrow$ & SSIM$\uparrow$ & LPIPS$\downarrow$ & \#G$\downarrow$
       & PSNR$\uparrow$ & SSIM$\uparrow$ & LPIPS$\downarrow$ & \#G$\downarrow$
       & PSNR$\uparrow$ & SSIM$\uparrow$ & LPIPS$\downarrow$ & \#G$\downarrow$ \\
\midrule
\multirow{6}{*}{\rotatebox[origin=c]{90}{DL3DV}}
& WorldMirror & \multicolumn{4}{c}{OOM} & \multicolumn{4}{c}{OOM} & \multicolumn{4}{c}{OOM} \\
& AnySplat & \multicolumn{4}{c}{OOM} & \multicolumn{4}{c}{OOM} & \multicolumn{4}{c}{OOM} \\
& ReCoSplat (official) & 13.62 & 0.234 & 0.647 & 8.78 & 13.67 & 0.237 & 0.654 & 13.17 & 13.70 & 0.241 & 0.660 & 17.56 \\
& ReCoSplat (adapted) & 15.95 & 0.288 & 0.599 & 8.78 & \secondbest{15.74} & 0.282 & 0.606 & 13.17 & \secondbest{15.48} & 0.276 & 0.612 & 17.56 \\
& OF3GS & \secondbest{16.29} & \secondbest{0.438} & \secondbest{0.489} & \secondbest{7.10} & 15.31 & \secondbest{0.404} & \secondbest{0.545} & \secondbest{10.45} & 14.83 & \secondbest{0.316} & \secondbest{0.574} & \secondbest{13.67} \\
& \textbf{LonGaussian} & \best{19.57} & \best{0.617} & \best{0.381} & \best{2.36} & \best{18.88} & \best{0.575} & \best{0.420} & \best{3.54} & \best{18.31} & \best{0.536} & \best{0.450} & \best{4.80} \\
\midrule
\multirow{6}{*}{\rotatebox[origin=c]{90}{RealEstate10K}}
& WorldMirror & \multicolumn{4}{c}{OOM} & \multicolumn{4}{c}{OOM} & \multicolumn{4}{c}{OOM} \\
& AnySplat & \multicolumn{4}{c}{OOM} & \multicolumn{4}{c}{OOM} & \multicolumn{4}{c}{OOM} \\
& ReCoSplat (official) & 13.66 & 0.358 & 0.609 & 8.78 & 13.72 & 0.359 & 0.618 & 13.17 & 13.67 & 0.358 & 0.627 & 17.56 \\
& ReCoSplat (adapted) & 16.34 & 0.431 & 0.589 & 8.78 & 15.85 & 0.421 & 0.601 & 13.17 & 15.34 & 0.412 & 0.612 & 17.56 \\
& OF3GS & \secondbest{18.47} & \secondbest{0.580} & \best{0.331} & \secondbest{6.04} & \secondbest{17.17} & \secondbest{0.518} & \secondbest{0.384} & \secondbest{8.74} & \secondbest{16.19} & \secondbest{0.473} & \secondbest{0.431} & \secondbest{11.37} \\
& \textbf{LonGaussian} & \best{19.73} & \best{0.636} & \secondbest{0.339} & \best{1.12} & \best{18.91} & \best{0.595} & \best{0.372} & \best{1.66} & \best{18.23} & \best{0.561} & \best{0.397} & \best{2.15} \\
\bottomrule
\end{tabular}
}
\end{table}

\begin{figure}[t]
\centering
    \includegraphics[width=\textwidth]{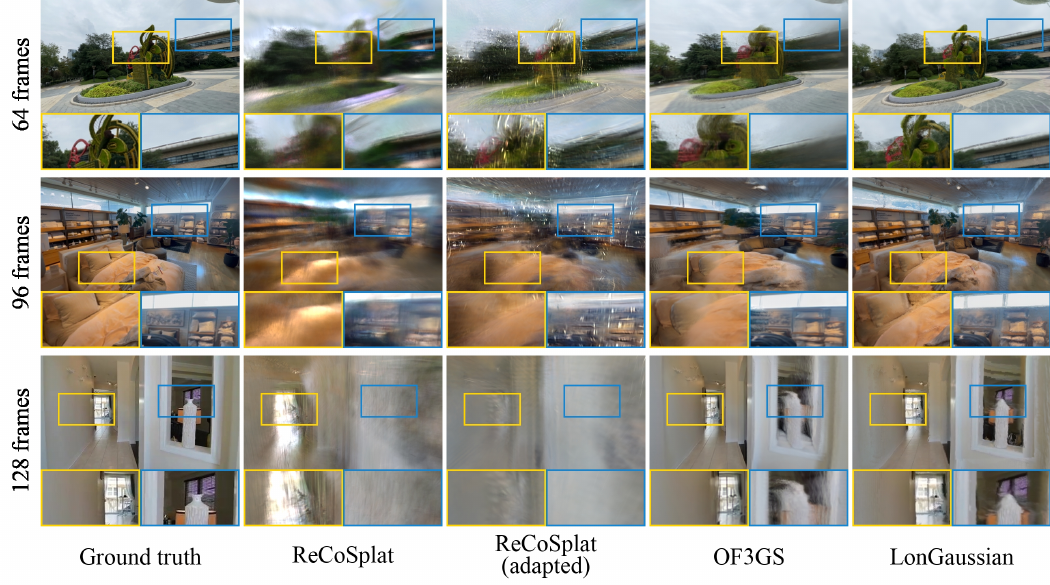}
\caption{Representative novel-view comparisons across sequence lengths. Colored boxes show that LonGaussian better preserves structure and detail; complete comparisons appear in the Appendix.}
\label{fig:main_qualitative}
\end{figure}

Table~\ref{tab:main_results} and Figure~\ref{fig:main_qualitative} compare reconstruction across sequence lengths. WorldMirror and AnySplat exceed the GPU memory budget even at the shortest evaluated length. Reconstructions from both ReCoSplat variants remain blurred or geometrically distorted, while OF3GS preserves coarse layout but loses fine detail and accumulates many Gaussians. LonGaussian instead achieves the best PSNR and SSIM at every length, produces sharper and more coherent renderings, and uses the fewest Gaussians. OF3GS retains a slight LPIPS advantage on RE10K at the shortest evaluated sequence length. As sequences grow, LonGaussian adds substantially fewer Gaussians than the streaming baselines on both datasets, supporting more compact reconstruction while preserving rendering quality. Videos on the project page provide side-by-side novel-view rendering comparisons for all evaluated methods.

\subsection{Long-Sequence Scalability}
\label{sec:long_scalability}

We evaluate six cumulative prefixes of the scene in Figure~\ref{fig:applications}, extending to 3,768 input frames. Unlike the complete-scene rendering in Table~\ref{tab:main_results}, image metrics here average 15 held-out views per prefix, each rendered from its owner chunk and up to three adjacent chunks on either side. This seven-chunk scope reduces interference from distant chunks misregistered by LoGeR's loop-closure errors. Gaussian count and construction resources measure the complete accumulated prefix.

\begin{table}[t]
\caption{Single-scene long-sequence scalability. Quality uses seven-chunk rendering (mean $\pm$ standard deviation across target views); \#G counts the complete filtered representation. Memory is peak allocated GPU memory during streaming construction. Time excludes model loading, metric rendering, and offline prefix filtering.}
\label{tab:long_scalability}
\centering
\small
\setlength{\tabcolsep}{3.5pt}
\begin{tabular}{rccccrr}
\toprule
Input $N$ & PSNR $\uparrow$ & SSIM $\uparrow$ & LPIPS $\downarrow$ & \#G (M) $\downarrow$ & Memory & Time \\
\midrule
126   & $20.33\pm1.87$ & $0.711\pm0.068$ & $0.404\pm0.061$ & 7.28   & 19.99 GB & 55 s \\
250   & $19.35\pm2.50$ & $0.681\pm0.083$ & $0.438\pm0.073$ & 15.04  & 20.05 GB & 101 s \\
507   & $18.40\pm2.83$ & $0.626\pm0.118$ & $0.485\pm0.097$ & 28.16  & 20.05 GB & 197 s \\
1,021 & $17.73\pm2.52$ & $0.624\pm0.094$ & $0.506\pm0.091$ & 57.72  & 20.05 GB & 392 s \\
2,040 & $18.24\pm2.31$ & $0.586\pm0.172$ & $0.530\pm0.111$ & 111.42 & 20.05 GB & 781 s \\
3,768 & $17.17\pm2.89$ & $0.556\pm0.125$ & $0.531\pm0.113$ & 208.29 & 20.05 GB & 1,445 s \\
\bottomrule
\end{tabular}
\end{table}

Table~\ref{tab:long_scalability} shows approximately linear growth in construction time and representation size, with peak construction memory remaining nearly constant on this sequence. Local rendering quality generally decreases as the prefix expands. Appendix~\ref{app:long_scalability} provides protocol details and scaling curves.

\subsection{Ablation Studies}

We ablate the components of LonGaussian on DL3DV-140 using 96 context frames and 13 target views per scene, with the same checkpoint, two-pass protocol, resolution, and split as Table~\ref{tab:main_results}. Dense accumulation retains pixel-aligned predictions, deterministic merging replaces the learned merger with fixed within-track aggregation, and each remaining variant removes one component.

\begin{table}[t]
\caption{Component ablation on DL3DV-140 at 96 frames. \#G is in millions; time is the per-scene reconstruction and evaluation time.}
\label{tab:ablation}
\centering
\small
\setlength{\tabcolsep}{5.0pt}
\begin{tabular}{lccccc}
\toprule
Configuration & PSNR$\uparrow$ & SSIM$\uparrow$ & LPIPS$\downarrow$ & \#G$\downarrow$ & Time (s)$\downarrow$ \\
\midrule
Dense accumulation & 18.48 & 0.567 & \textbf{0.416} & 10.02 & 155.8 \\
Deterministic merging & 18.63 & 0.569 & 0.432 & 3.60 & 138.6 \\
w/o dense complement & 17.53 & 0.550 & 0.424 & \textbf{0.95} & 124.8 \\
w/o depth filtering & 17.36 & 0.482 & 0.486 & 4.86 & 121.5 \\
w/o far-field shell & 18.59 & 0.569 & 0.425 & 3.71 & \textbf{103.4} \\
\textbf{Full model} & \textbf{18.88} & \textbf{0.575} & 0.420 & 3.54 & 132.6 \\
\bottomrule
\end{tabular}
\end{table}

\begin{figure}[htbp]
    \centering
    \includegraphics[width=0.90\textwidth]{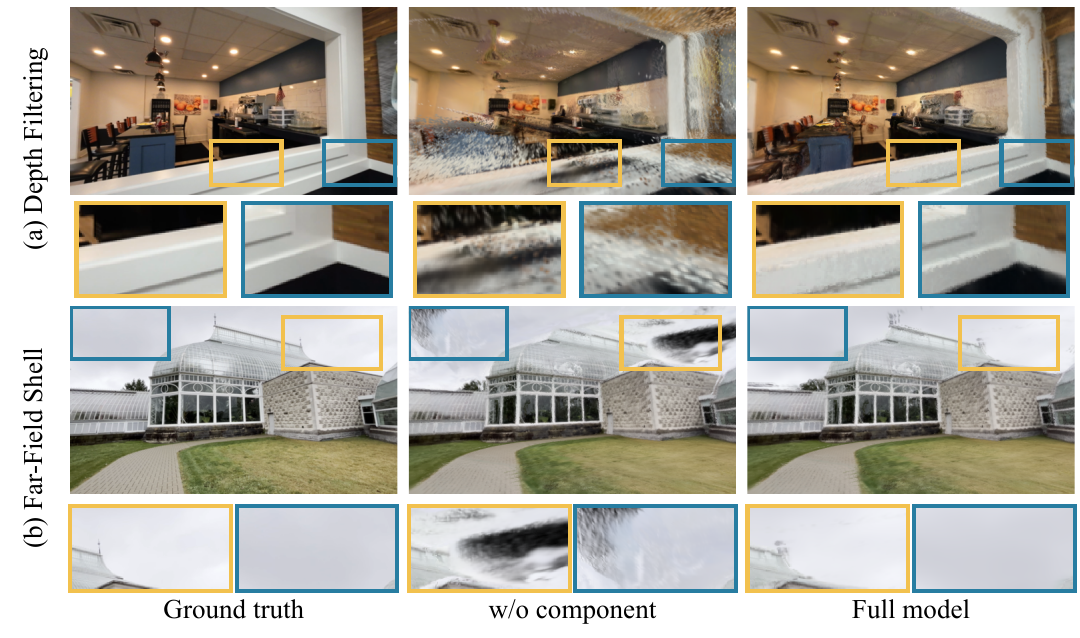}
        \caption{Qualitative artifact ablations. Rendered-depth filtering removes cross-chunk floaters, while far-field modeling avoids unstable sky geometry; colored boxes mark enlarged regions.}
    \label{fig:artifact_handling_qualitative}
\end{figure}

Table~\ref{tab:ablation} shows that the full model substantially reduces Gaussian count while improving PSNR and SSIM, with a slight LPIPS trade-off relative to dense accumulation. Learned refinement improves rendering quality over deterministic merging at a similar representation size. Removing the dense complement yields a smaller representation but degrades reconstruction quality, demonstrating the importance of preserving content without reliable track support. Depth filtering provides the largest PSNR gain by suppressing cross-chunk occluders, while the far-field shell addresses unstable sky geometry (Figure~\ref{fig:artifact_handling_qualitative}). Disabling either reduces reconstruction and evaluation time but lowers rendering quality: despite retaining more Gaussians, these variants run faster because they omit processing stages. Thus, Gaussian count alone does not determine total runtime.

\section{Discussion and Limitations}
\label{sec:limitations}

LonGaussian assumes static scenes and reliable tracks. Motion, geometric errors, and fragmented tracks may cause reconstruction errors, while within-chunk tracking may duplicate surfaces revisited after long gaps. Depth voting and shell construction add overhead. On long-range revisits, LoGeR pose errors can misalign accumulated Gaussians and cause global rendering ghosting. The seven-chunk metrics reduce interference from distant misregistered chunks and therefore do not measure full-scene global rendering fidelity. Multi-scene benchmarks cover at most 128 frames; Section~\ref{sec:long_scalability} extends a single-scene quantitative case study to 3,768 frames and a qualitative example to 3,773 frames. Future work could address dynamic scenes, cross-chunk association, uncertainty-aware filtering, and adaptive far-field representations.

\section{Conclusion}
\label{sec:conclusion}

LonGaussian reconstructs compact, globally renderable Gaussian scenes from long sequences by combining LoGeR's chunk-wise hybrid memory with track-guided consolidation and artifact filtering. Tracker-derived correspondences merge repeated surface observations, while rendered-depth filtering and far-field modeling suppress cross-chunk and sky artifacts. Experiments on DL3DV and RealEstate10K show that LonGaussian improves rendering quality while using substantially fewer Gaussians than feed-forward baselines, connecting long-context geometry to efficient renderable representations.

\subsection*{AI Use Statement}

In this work, we used generative AI tools to assist with the
implementation of selected components of the proposed method
and the translation of manuscript content into English.
We also used these tools for literature search and
summarization, code debugging, and language refinement.

Generative AI tools were not used to develop the research
methodology, design experiments, generate or process datasets,
formulate mathematical claims or proofs, create scientific
figures, or interpret experimental results.

All AI-assisted work was reviewed by the authors, and the
authors take full responsibility for the final content of
this paper, including all text, scientific claims, code,
and other artifacts produced with the assistance of
generative AI.

\subsection*{Ethics Statement}

This work studies the reconstruction of static 3D scenes from image sequences and does not involve human-subject experiments or the collection of new personal data. We use existing benchmark datasets and pretrained models for research purposes and follow their applicable terms of use. As with other scene-reconstruction systems, LonGaussian may reproduce identifiable or private visual content present in its inputs and could be misused for unauthorized reconstruction or surveillance. Its deployment should therefore respect privacy, consent, copyright, dataset licenses, and applicable regulations. The method may also inherit geographic and visual biases from its training data, and we do not evaluate it for safety-critical deployment.

\subsection*{Reproducibility Statement}

Sections~\ref{sec:method} and~\ref{sec:experiments} describe the model components, two-stage training procedure, objectives, datasets, evaluation protocol, and principal implementation settings. The appendix provides additional architecture and optimization details, hyperparameters, data-processing procedures, sequence construction rules, baseline adaptations, and evaluation settings. Side-by-side novel-view rendering comparison videos are available on the project page, where code will also be released. These materials are intended to support reproduction of the reported results.

\clearpage
\appendix

\section{Architecture and Consolidation Details}
\label{app:implementation}

\subsection{Dense Gaussian Prediction Head}

The Gaussian branch follows a DPT-style dense prediction design. It reassembles four intermediate decoder features at complementary spatial scales, fuses them from coarse to fine, and produces a 256-channel full-resolution feature map. Separate two-layer convolutional heads predict a 3D offset, anisotropic scale, unit-quaternion rotation, opacity, and degree-2 spherical-harmonic coefficients for every context pixel. The heads therefore predict 3, 3, 4, 1, and 27 values per pixel for these attributes, respectively. The opacity-head bias is initialized to $-2$, while the final offset layer is initialized close to zero, preventing opaque or strongly displaced primitives at the start of training.

\subsection{Track Validation and Ownership}

AllTracker is run independently in each LoGeR chunk with four recurrent inference iterations and a confidence threshold $\tau_{\mathrm{trk}}=0.1$. Tracked coordinates are rounded to the dense Gaussian grid. When multiple tracks map to the same frame--pixel location, only the observation with the highest opacity-weighted visibility is retained, so a dense prediction cannot be consumed by more than one representative. A track is eligible for consolidation when it contains at least two retained observations, has a mean opacity of at least $\tau_\alpha=0.05$, and has a root-mean-square distance of its 3D centers from their arithmetic mean of at most $\tau_{\mathrm{geo}}=0.3$. The threshold $\tau_{\mathrm{geo}}$ is an absolute distance in the backbone reconstruction coordinates, not a calibrated distance in meters or a scene-normalized quantity. Intrinsics calibration preserves predicted depth, and SE(3) alignment does not change this scale. We additionally restrict observations to an adaptive neighborhood around the query-frame anchor. Its radius is
\begin{equation}
r_{\mathrm{trk}}
=\operatorname{clip}\!\left(6d_{\mathrm{1NN}},0.02,0.5\right),
\label{eq:app_track_radius}
\end{equation}
where $d_{\mathrm{1NN}}$ is the median nearest-neighbor distance estimated from at most 5,000 points in the chunk. This neighborhood gate adapts to local point spacing within the clipping bounds; it does not make the separate absolute threshold $\tau_{\mathrm{geo}}$ scale invariant. We keep $\tau_{\mathrm{geo}}$ fixed across scenes and do not independently evaluate its sensitivity.

\subsection{Learned Track Merger}

For every accepted track, the base center, scale, and SH coefficients are opacity-weighted averages of its observations. The base opacity is obtained by alpha compositing, and the base rotation is computed by weighted quaternion averaging. The pooled descriptor defined in Section~\ref{sec:track_consolidation} has 69 dimensions and is processed by a two-layer MLP with hidden dimension 128, LayerNorm, and GELU activations. Four zero-initialized output layers predict bounded corrections:
\begin{equation}
\begin{aligned}
\hat{\boldsymbol\mu}_m &= \bar{\boldsymbol\mu}_m
    +0.2\tanh(\Delta\boldsymbol\mu_m), \\
\hat{\mathbf s}_m &= \bar{\mathbf s}_m
    \odot\exp\!\left(0.5\tanh(\Delta\mathbf s_m)\right),
\end{aligned}
\label{eq:app_merger_residuals}
\end{equation}
while the opacity correction is applied in logit space and clipped to $[-4,4]$, and the SH correction is scaled by $0.25\tanh(\cdot)$.

\section{Scene-Level Artifact Handling}
\label{app:artifact_handling}

\subsection{Rendered-Depth Consistency Filtering}

For each chunk, we render its Gaussian predictions in its contributing context views and use the resulting depth and opacity as local references. Gaussians originating from other chunks are projected into these views and compared with the median depth in a $3\times3$ neighborhood. A comparison is ignored when fewer than five neighbors have sufficient opacity or when their relative depth spread exceeds $0.15$. Otherwise, the comparison contributes an occlusion vote when the projected Gaussian lies at least $1\%$ in front of the reference surface and an agreement vote when the relative depth difference is at most $1\%$. A Gaussian is removed only if it receives at least one occlusion vote, occlusion votes constitute at least $30\%$ of all reliable comparisons, and they also constitute at least $30\%$ of the combined occlusion and agreement votes. The reference-opacity threshold is $0.5$. A chunk never votes against its own predictions. These conservative ownership and support tests prevent a single uncertain depth sample from deleting locally consistent geometry.

\subsection{Filtering-Threshold Selection}
\label{app:depth_filter_thresholds}

We select the reported filtering decision thresholds exclusively on a subset of the DL3DV training split, disjoint from DL3DV-140. Specifically, we randomly sample 20 training scenes and reconstruct each using 96 context and 13 held-out target views. We evaluate nine predefined joint threshold configurations rather than an exhaustive Cartesian grid. Candidate filters are replayed on the same cached, unfiltered reconstructions, so their differences arise only from thresholding rather than separate model training or reconstruction. We rank candidates by scene-equal mean target-view PSNR. The selected setting uses a $1\%$ relative depth gap, at least one occlusion vote, a $30\%$ reliable-view ratio, and a $30\%$ decision ratio. It achieves the highest training-subset PSNR among the candidates, improves all 20 scenes over the matched no-filter reconstruction, and is fixed before evaluation on the test sets. The DL3DV-140 ablation in Table~\ref{tab:ablation} evaluates this fixed filter and is not used for threshold selection. Across the tested joint configurations, filtering improves mean PSNR, SSIM, and LPIPS over the matched unfiltered reconstruction. Nearby candidates yield similar mean PSNR, suggesting that the benefit is not confined to the selected setting. This comparison varies depth gaps and voting criteria jointly; it does not isolate each parameter's effect or establish robustness outside the tested configurations.

\begin{table}[h]
\caption{Rendered-depth threshold selection on 20 DL3DV training scenes. Setting tuples denote (relative gap, minimum votes, reliable-view ratio, decision ratio). Metrics are averaged equally across scenes; $\Delta$PSNR is paired against the same reconstruction without filtering.}
\label{tab:depth_filter_threshold_selection}
\centering
\small
\setlength{\tabcolsep}{4.5pt}
\begin{tabular}{lcccc}
\toprule
Setting & PSNR $\uparrow$ & SSIM $\uparrow$ & LPIPS $\downarrow$ & $\Delta$PSNR \\
\midrule
$(0.10,2,0.50,0.65)$ & 17.418 & 0.514 & 0.464 & $+0.550$ \\
$(0.05,2,0.50,0.65)$ & 17.631 & 0.524 & 0.457 & $+0.763$ \\
$(0.05,1,0.50,0.50)$ & 17.688 & 0.527 & 0.456 & $+0.820$ \\
$(0.05,1,0.40,0.40)$ & 17.795 & 0.534 & 0.452 & $+0.926$ \\
$(0.02,2,0.50,0.50)$ & 17.967 & 0.541 & 0.449 & $+1.099$ \\
$(0.02,1,0.40,0.40)$ & 18.137 & 0.551 & 0.443 & $+1.269$ \\
$(0.02,1,0.30,0.30)$ & 18.271 & 0.560 & 0.438 & $+1.403$ \\
$(0.01,1,0.40,0.40)$ & 18.279 & 0.559 & 0.440 & $+1.411$ \\
Selected $(0.01,1,0.30,0.30)$ & \textbf{18.356} & \textbf{0.566} & \textbf{0.435} & \textbf{$+1.488$} \\
No filtering & 16.868 & 0.484 & 0.487 & --- \\
\bottomrule
\end{tabular}
\end{table}

\subsection{Far-Field Gaussian Shell}

We use the public \texttt{skyseg.onnx} checkpoint hosted by Jianyuan Wang.\footnote{\url{https://huggingface.co/JianyuanWang/skyseg}} Each RGB frame is resized to $320\times320$, normalized with ImageNet statistics, and processed by the fixed ONNX model. Its score map is min--max normalized per image to unsigned values in $[0,255]$ and bilinearly resized to the input resolution; pixels with scores greater than 160 are classified as sky. The resulting mask first suppresses finite-depth Gaussian predictions. We define the shell center as the coordinate-wise median of the input camera centers and estimate the scene radius as the 90th percentile of the distances from scene Gaussians to this center. The shell radius is $r_{\mathrm{sky}}=50r_{\mathrm{scene}}$. Sky pixels are sampled with stride 2, converted to world-space rays, and quantized into $512\times256$ azimuth--elevation bins. Each occupied bin produces one isotropic Gaussian with opacity 0.98. Its direction and RGB color are averaged from the observations in the bin; RGB is represented by degree-zero SH, while the remaining coefficients are zero. The Gaussian scale equals the shell radius times the angular bin size, with an overlap factor of 1.0.

\section{Training Details}
\label{app:training_details}

\subsection{Two-Stage Optimization}

All stages keep LoGeR and AllTracker frozen and use a global batch size of four. Stage~I trains the Gaussian decoder and attribute heads for one complete pass over the 8,590 eligible scenes in the DL3DV training split after excluding all scenes that overlap with DL3DV-140; each sample is a dense 16-frame clip. Stage~II initializes from Stage~I and freezes the feature decoder. In Stage~IIa, we additionally freeze the Gaussian attribute head and train only the learned merger on DL3DV for 900 optimizer steps. Stage~IIb initializes from the Stage~IIa checkpoint and performs a 300-step low-learning-rate adaptation with a balanced 1:1 schedule over 150 sequences from the official RE10K training split and 118 DL3DV training scenes; neither adaptation set contains an evaluation identity. In both substages, we sample uniformly from four sequence-length buckets: a 16-frame sparse sequence sampled at interval 8 and contiguous sequences of 64, 96, and 128 frames. During Stage~IIb, the merger is updated in every step, whereas the attribute head is updated only for the 16-frame bucket; this corresponds to approximately 75 head updates under uniform sampling. At most three context views and three held-out target views are rendered in each update.

We use AdamW with $\beta_1=0.9$, $\beta_2=0.999$, $\epsilon=10^{-8}$, and weight decay $0.01$, without learning-rate scheduling, and clip the gradient norm at 1.0. Training uses four GPUs with one sequence per GPU. The Stage-I learning rate is $10^{-5}$. In Stage~IIa, the merger learning rate is $1.25\times10^{-5}$. Stage~IIb retains this merger learning rate and uses $2\times10^{-6}$ for the attribute head. The image-loss weights are $\lambda_{\mathrm{rgb}}=1$, $\lambda_{\mathrm{lpips}}=0.25$, $\lambda_{\mathrm{ssim}}=0.15$, and $\lambda_{\mathrm{tgt}}=1$. Stage~I additionally uses $\lambda_{\mathrm{aniso}}^{(1)}=0.1$ with anisotropy threshold 4 and $\lambda_{\mathrm{off}}^{(1)}=0.01$. Because Stage~II supervises only the consolidated rendering branch, $\lambda_{\mathrm{aniso}}^{(2)}=\lambda_{\mathrm{off}}^{(2)}=0$. All experiments use random seed 0.

For dense predictions indexed by $i=1,\ldots,M$, with scale vector $\mathbf s_i$ and predicted center offset $\Delta\boldsymbol\mu_i$, the two regularizers are
\begin{equation}
\mathcal L_{\mathrm{aniso}}
=\frac{1}{M}\sum_{i=1}^{M}
\max\!\left(0,
\frac{\max_j s_{ij}}{\min_j s_{ij}+10^{-6}}-\tau_{\mathrm{aniso}}\right),
\qquad
\mathcal L_{\mathrm{off}}
=\frac{1}{M}\sum_{i=1}^{M}\left\|\Delta\boldsymbol\mu_i\right\|_2,
\label{eq:regularization_losses}
\end{equation}
where $\tau_{\mathrm{aniso}}=4$. These terms discourage needle-like Gaussians and large displacements from the LoGeR point-map predictions, respectively.

\begin{table}[h]
\caption{Principal architecture, training, and inference hyperparameters.}
\label{tab:hyperparameters}
\centering
\small
\setlength{\tabcolsep}{5pt}
\begin{tabular}{lc}
\toprule
Hyperparameter & Value \\
\midrule
Image resolution & $280\times490$ \\
LoGeR chunk length / overlap $(L,O)$ & $(12,3)$ \\
Gaussian SH degree & 2 \\
Merger hidden dimension & 128 \\
AllTracker confidence threshold $\tau_{\mathrm{trk}}$ & 0.1 \\
Minimum track length $n_{\min}$ & 2 \\
Minimum mean opacity $\tau_\alpha$ & 0.05 \\
Maximum RMS center deviation $\tau_{\mathrm{geo}}$ & 0.3 \\
Head / merger learning rate & $2\times10^{-6}$ / $1.25\times10^{-5}$ \\
Stage-I / Stage-II schedule & 8,590 scenes (one pass) / $900{+}300$ steps \\
AdamW weight decay & $0.01$ \\
Context / target renders per update & at most 3 / 3 \\
Depth gap / visibility ratio & 0.01 / 0.30 \\
Depth neighborhood / minimum support & 1 / 5 \\
Maximum local depth spread & 0.15 \\
Sky radius factor / sampling stride & 50 / 2 \\
Sky angular bins & $512\times256$ \\
\bottomrule
\end{tabular}
\end{table}

\section{Evaluation Protocol and Baselines}
\label{app:evaluation_protocol}

\subsection{Dataset Preparation and View Sampling}

DL3DV uses its downsampled images, while RealEstate10K frames are center-cropped to the target aspect ratio and resized to $280\times490$ with corresponding updates to the intrinsics. For evaluation with $C$ context views, every eighth frame in a contiguous source prefix is held out as a target. The prefix therefore contains $C+\lfloor C/7\rfloor$ frames: the 64-, 96-, and 128-frame context settings contain 9, 13, and 18 target views, respectively. Scene lists and frame indices are fixed before evaluation and shared across methods whenever their interfaces permit.

\subsection{Two-Pass Unposed Evaluation}
\label{app:two_pass_evaluation}

Our unposed evaluation separates camera estimation from scene construction. The first pass processes the complete packed sequence to estimate a camera trajectory for both context and target views. The second pass reinitializes the TTT fast-weight states and SWA cache, receives only context images, and constructs the Gaussian scene with tracking, consolidation, filtering, and shell generation. The first-pass camera trajectory is aligned to the second-pass scene coordinate system using the shared context cameras, after which the aligned held-out target poses are used for rendering. Consequently, target images provide camera-pose observations in the first pass but never contribute Gaussian primitives, tracking correspondences, appearance attributes, depth-filter references, or sky-shell samples. This protocol evaluates renderable reconstruction from unposed context images without using target appearance in the reconstructed scene.

All LonGaussian experiments use SE(3), enable LoGeR's TTT and sliding-window attention, and calibrate local point maps with the dataset intrinsics. We use orientation-aware two-pass alignment and disable Sim(3) rescaling. PSNR, SSIM, and LPIPS-VGG are computed with the same implementation for all methods. The Gaussian count is measured after all scene construction and filtering steps specific to each method. ``OOM'' denotes failure to complete a method's native inference on a 24-GB GPU; we do not introduce method-specific chunking solely to obtain a result.

\subsection{Baseline Configuration}

For the baselines with numerical results in Table~\ref{tab:main_results}, target poses are predicted independently by each method rather than shared with LonGaussian or taken directly from the dataset. Both ReCoSplat checkpoints use a first pass over context and target images for pose estimation and a separate context-only pass for Gaussian construction. Their target translations are rescaled using the predicted context-camera scale ratio between passes, with a shared first-camera coordinate frame; the prescribed first-chunk scale matching is retained. OF3GS likewise predicts poses from context and target images, constructs Gaussians from context images, and applies its native context-based translation-scale matching before target rendering. These image metrics therefore reflect both scene reconstruction and each method's pose-estimation errors, and are not equivalent to evaluation with fixed ground-truth target poses. WorldMirror and AnySplat yield only OOM entries at the reported lengths.

WorldMirror~\citep{liu2026worldmirror} and AnySplat use their native joint-view inference. OF3GS uses its native streaming construction. For ReCoSplat, we report both the released low-resolution checkpoint and a locally adapted checkpoint trained at $280\times490$ on a single 24-GB GPU. Starting from the released weights, adaptation comprises 1,000 updates of the Gaussian decoder and output head, 2,000 updates that additionally train the recurrent-guidance projections, and 1,000 mixed-context updates over 32, 64, 96, and 128 frames. All three phases use a learning rate of $10^{-5}$ and approximately 975 DL3DV training scenes disjoint from DL3DV-140; the backbone and camera and point prediction paths remain frozen. The final checkpoint is selected using non-RE10K DL3DV development evidence before RE10K evaluation. This procedure preserves the released architecture and inference protocol but is a local resolution adaptation rather than an official ReCoSplat reproduction. All available methods use the same prepared scene and target-view lists.

\subsection{Single-Scene Long-Sequence Scalability}
\label{app:long_scalability}

\begin{figure}[t]
    \centering
    \includegraphics[width=0.98\textwidth]{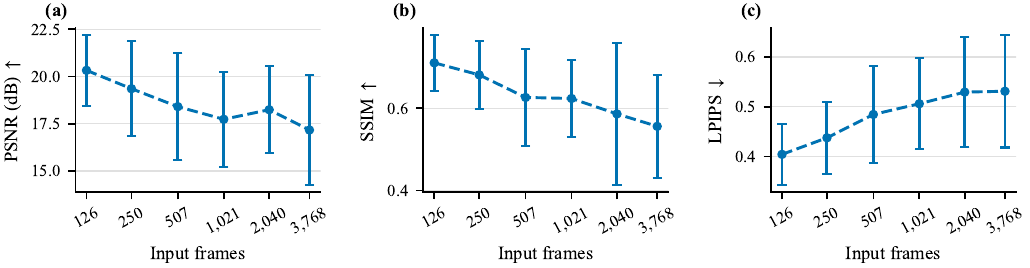}
    \vspace{2pt}
    \includegraphics[width=0.98\textwidth]{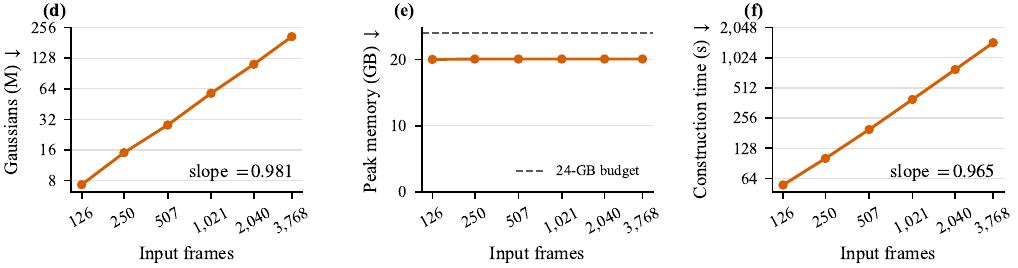}
    \caption{Single-scene long-sequence scaling. Local rendering quality remains usable as the sequence grows (a--c), while complete representation size and construction time grow approximately linearly and peak memory remains bounded (d--f). Error bars show variation across held-out views.}
    \label{fig:long_scalability}
\end{figure}

The prefixes in Table~\ref{tab:long_scalability} preserve complete chunk boundaries; the final five frames are included only in the qualitative 3,773-frame reconstruction in Figure~\ref{fig:applications}. We process the sequence once at $280\times490$ with $L=12$ and $O=3$ and record these six streaming snapshots. Context frames follow the same one-in-eight held-out split as the main evaluation. The 15 target views per prefix are sampled uniformly from the held-out views and use the seven-chunk rendering scope defined in Section~\ref{sec:long_scalability}. Filtering is recomputed within each prefix, and the rendered representation contains the prefix-filtered compact component and dense complement with an $8\%$ compact-depth surface-culling margin. Because the snapshots share the full run's first-pass pose trajectory, this experiment measures scaling under a fixed trajectory rather than strictly causal prefix-only camera estimation.

Table~\ref{tab:long_scalability} reports the combined quality and resource measurements. For its six prefixes, the context counts are 111, 219, 444, 894, 1,785, and 3,297, corresponding to 12, 24, 49, 99, 198, and 366 chunks. Construction time fits $T\propto N^{0.965}$ ($R^2=0.999$), while representation size follows $G\propto C^{0.981}$ ($R^2=1.000$). Time per input frame and Gaussian count per context view remain approximately stable. Figure~\ref{fig:long_scalability} visualizes these trends and the target-view variation. These single-scene fits do not establish cross-scene scaling rates.

Because each longer prefix introduces new scene regions and different held-out views, this diagnostic combines increased content difficulty with local pose and alignment error and should not be interpreted as a controlled fixed-content degradation curve.

\section{Additional Analysis and Failure Cases}
\label{app:additional_analysis}

\begin{figure}[htbp]
    \centering
    \includegraphics[width=\textwidth]{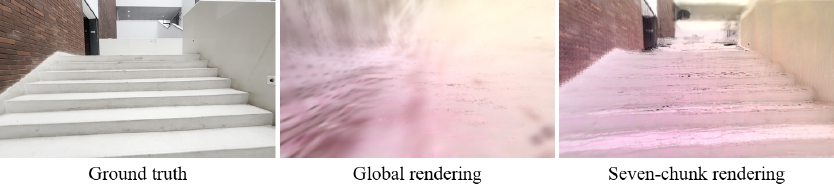}
    \caption{Long-range revisit failure in the ultra-long sequence. Rendering the complete accumulated scene produces severe ghosting, whereas restricting the same target view to seven neighboring chunks recovers much of the scene structure. The improvement is consistent with interference from distant misregistered chunks; residual local artifacts remain.}
    \label{fig:loop_revisit_failure}
\end{figure}

The Gaussian counts in Table~\ref{tab:main_results} include track representatives, uncovered dense predictions, and far-field shell primitives. Runtime also includes frozen tracking, two LoGeR passes, depth voting, and sky segmentation.

Tracks on moving objects may link observations that do not share a single 3D location, while textureless, reflective, or repeatedly occluded surfaces may fragment tracks and leave duplicated dense predictions. Geometric errors that are consistent within a LoGeR chunk can pass the track test. The depth filter is deliberately conservative but can still remove thin foreground structures when their local rendered depth is unstable, and an erroneous sky mask may replace finite scene content with shell Gaussians.

\begin{figure}[p]
    \centering
    \includegraphics[width=\textwidth]{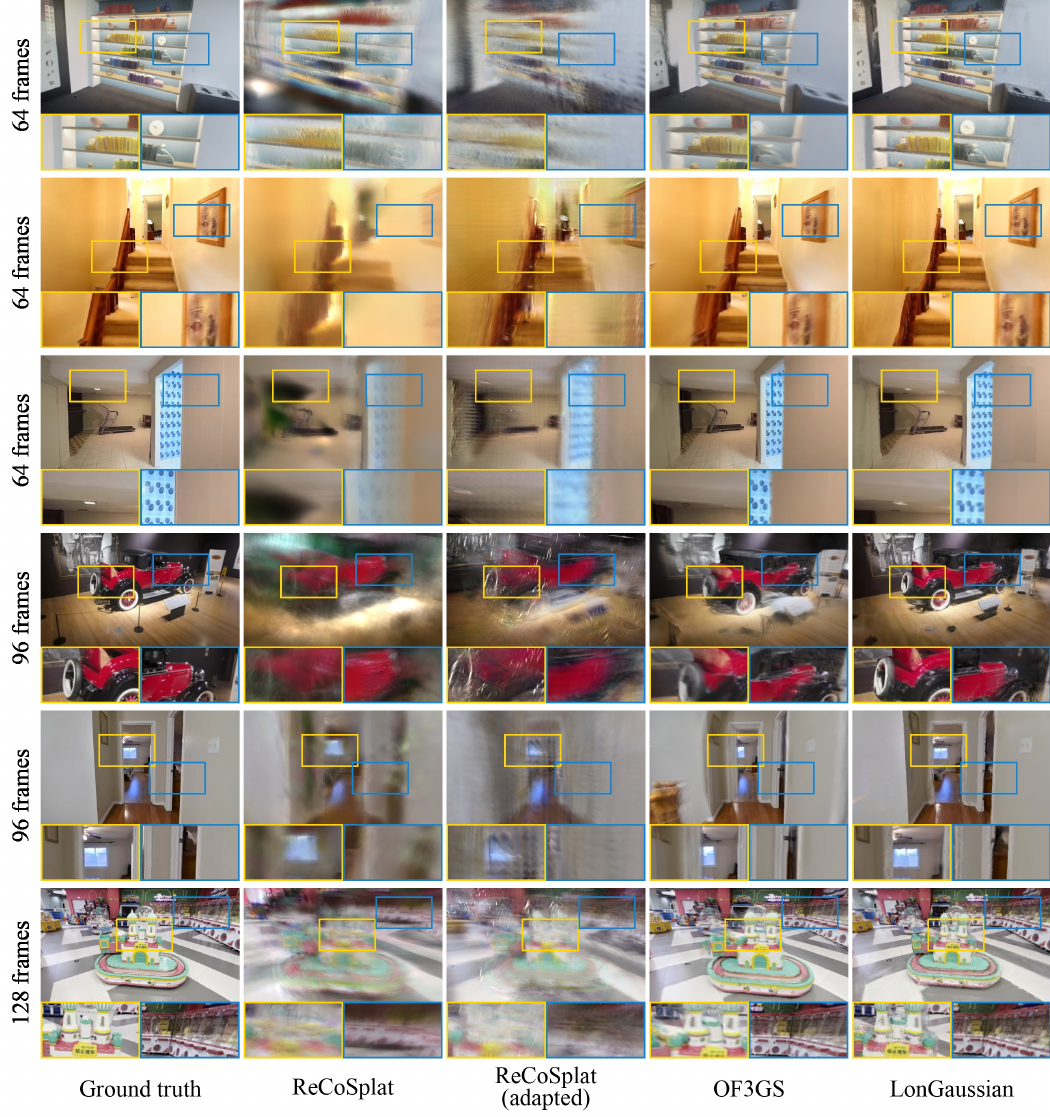}
    \caption{Additional novel-view comparisons on DL3DV and RealEstate10K. LonGaussian consistently preserves sharper structure and finer detail across scenes and sequence lengths.}
    \label{fig:main_qualitative_full}
\end{figure}

\end{document}

%% file: math_commands.tex
\usepackage{amsmath,amsfonts,bm}

\def\1{\bm{1}}

\DeclareMathAlphabet{\mathsfit}{\encodingdefault}{\sfdefault}{m}{sl}
\SetMathAlphabet{\mathsfit}{bold}{\encodingdefault}{\sfdefault}{bx}{n}